\documentclass[11pt]{article}
\usepackage{nodalida2025}
\usepackage{times}  
\usepackage[hyphens,spaces,obeyspaces]{url}
\usepackage{latexsym}
\usepackage[table,dvipsnames]{xcolor}
\usepackage[utf8]{inputenc} 
\usepackage[T1]{fontenc}
\usepackage{xcolor}
\definecolor{darkblue}{rgb}{0, 0, 0.5}
\usepackage{hyperref}
\hypersetup{
           breaklinks=true,   
           colorlinks=true,   
           pdfusetitle=true,  
           citecolor=darkblue, 
           linkcolor=darkblue, 
           urlcolor=darkblue
        }
\usepackage{amsfonts}       
\usepackage{nicefrac}       
\usepackage{microtype}
\usepackage{soul}
\usepackage{inconsolata}

\usepackage{times,latexsym}
\usepackage{url}
\usepackage{pgf}
\usepackage{booktabs}

\usepackage{expex}
\usepackage[linguistics]{forest}

\usepackage{extdash} 

\definecolor{Gray}{gray}{0.9}
\definecolor{cb-blue-green} {RGB}{ 0,  073,  073}
\definecolor{cb-green-sea}  {RGB}{ 0, 146, 146}
\definecolor{cb-rose}       {RGB}{255, 109, 182}
\definecolor{cb-salmon-pink}{RGB}{255, 182, 119}
\definecolor{cb-purple}     {RGB}{ 73,   0, 146}
\definecolor{cb-blue}       {RGB}{ 0, 109, 219}
\definecolor{cb-lilac}      {RGB}{182, 109, 255}
\definecolor{cb-blue-sky}   {RGB}{109, 182, 255}
\definecolor{cb-blue-light} {RGB}{182, 219, 255}
\definecolor{cb-burgundy}   {RGB}{146,   0,   0}
\definecolor{cb-brown}      {RGB}{146,  73,   0}
\definecolor{cb-clay}       {RGB}{219, 209,   0}
\definecolor{cb-green-lime} {RGB}{ 36, 255,  36}
\definecolor{cb-yellow}     {RGB}{255, 255, 109}
\definecolor{cb-grey}       {RGB}{233, 233, 233}

\usepackage{hhline}
\usepackage{amsmath}
\usepackage{amssymb}
\usepackage{multirow}
\usepackage{arydshln}
\usepackage[inline]{enumitem}
\usepackage{multicol}
\usepackage{xspace}
\usepackage{wasysym}
\usepackage{hyperref}

\usepackage{graphicx,scalerel} 
\graphicspath{{./images/}} 
\usepackage{adjustbox}
\usepackage{graphicx}

\usepackage{tikz}
\usepackage{collcell}
\usepackage{amssymb}
\usepackage{pifont}
\newcommand{\cmark}{\textcolor{cb-blue-green}{\ding{51}}}%
\newcommand{\xmark}{\textcolor{cb-burgundy}{\ding{55}}}%

\newcommand\rotation{0}
\newcommand*{\MinNumber}{0.0}%
\newcommand*{\MidNumber}{20.0} %
\newcommand*{\MaxNumber}{100.0}%

\newcommand{\ApplyGradient}[1]{%
        \ifdim #1 pt > \MidNumber pt
            \pgfmathsetmacro{\PercentColor}{max(min(100.0*(#1 - \MidNumber)/(\MaxNumber-\MidNumber),100.0),0.00)} %
            \hspace{-0.33em}\colorbox{SeaGreen!\PercentColor!Goldenrod!50}{#1}
        \else
            \pgfmathsetmacro{\PercentColor}{max(min(100.0*(\MidNumber - #1)/(\MidNumber-\MinNumber),100.0),0.00)} %
            \hspace{-0.33em}\colorbox{Red!\PercentColor!Goldenrod!50}{#1}
        \fi
}

\newcolumntype{R}{>{\collectcell\ApplyGradient}c<{\endcollectcell}}

\RequirePackage{etoolbox}
\RequirePackage{xcolor}
\definecolor{Gray}{gray}{0.9}
\definecolor{cb-blue}       {RGB}{ 0, 109, 219}

\aclfinalcopy 

\title{A Collection of Question Answering Datasets for Norwegian}

\author{Vladislav Mikhailov \hspace{0.7em} Petter Mæhlum \hspace{0.7em} Victoria Ovedie Chruickshank Langø\\ 
\textbf{Erik Velldal} \hspace{0.7em} \textbf{Lilja Øvrelid}\\
University of Oslo \\
\small{
    \textbf{Correspondence:} \href{mailto:vladism@ifi.uio.no}{\texttt{vladism@ifi.uio.no}}
}}
\date{}

\begin{document}
\maketitle
\begin{abstract}
This paper introduces a new suite of question answering datasets for Norwegian; NorOpenBookQA, NorCommonSenseQA, NorTruthfulQA, and NRK-Quiz-QA. The data covers a wide range of skills and knowledge domains, including world knowledge, commonsense reasoning, truthfulness, and knowledge about Norway. Covering both of the written standards of Norwegian -- Bokmål and Nynorsk -- our datasets comprise over 10k question-answer pairs, created by native speakers.  
We detail our dataset creation approach and present the results of evaluating 11 language models (LMs) in zero- and few-shot regimes. Most LMs perform better in Bokmål than Nynorsk, struggle most with commonsense reasoning, and are often untruthful in generating answers to questions. All our datasets and annotation materials are publicly available.

\end{abstract}

\begin{table*}[th!]
\centering
\scriptsize
\resizebox{\textwidth}{!}{
    \begin{tabular}{lrrrrr}
    \toprule
    & \textbf{NB / NN} &  \textbf{Size} & \textbf{Answer Evidence} & \textbf{Answer Format} & \textbf{Method} \\
    \midrule
    \textbf{NO-BoolQ} &  \cmark / \xmark & 12.7k & Context document & Yes/No & Machine translation \\
    \textbf{NorQuAD} & \cmark / \xmark & 4.7k  & Context document & Extractive & Human annotation \\
    \textbf{NO-Multi-QA-Sum} & \cmark / \xmark & 2.7k & Context document & Free form & \begin{tabular}{@{}c@{}} Model annotation \\ Human annotation \end{tabular}  \\
    \textbf{Belebele} & \cmark / \xmark & 900 & Context document &  Multiple choice & Human translation  \\
    \textbf{MKQA} &  \cmark / \xmark & 6.7k & World knowledge & Free form & Human translation \\
    \vspace{-1em} \\
    \cdashline{1-6}
    \vspace{-1em}\\
    \textbf{NRK-Quiz-QA} & \cmark / \cmark & 4.9k & \begin{tabular}{@{}c@{}} Norwegian-specific  \\ \& world knowledge \end{tabular} & Multiple choice & Human annotation  \\
    \vspace{-1em} \\
    \cdashline{1-6}
    \vspace{-1em}\\
    \textbf{NorOpenBookQA} & \cmark / \cmark & 3.5k & World knowledge & Multiple choice & \begin{tabular}{@{}c@{}} Human translation \\ Human annotation \end{tabular} \\
    \vspace{-1em} \\
    \cdashline{1-6}
    \vspace{-1em}\\
    \textbf{NorCommonSenseQA} & \cmark / \cmark & 1.1k & Common sense & Multiple choice & \begin{tabular}{@{}c@{}} Human translation \\ Human annotation \end{tabular} \\
    \vspace{-1em} \\
    \cdashline{1-6}
    \vspace{-1em}\\
    \multirow{2}{*}{\textbf{NorTruthfulQA}} & \cmark / \cmark & 545 & \multirow{2}{*}{Truthfulness} & Multiple choice & Human translation \\
    & \cmark / \cmark & 471 & & Free form & Human annotation \\
    \bottomrule
    \end{tabular}
}
\caption{Comparison of question answering resources for Norwegian: Belebele \cite{bandarkar-etal-2024-belebele}, NorQuAD \cite{ivanova-etal-2023-norquad}, MKQA \cite{longpre-etal-2021-mkqa}, NO-BoolQ \& NO-Multi-QA-Sum \cite{liu-etal-2024-nlebench}, and NRK-Quiz-QA, NorOpenBookQA, NorCommonSenseQA, and NorTruthfulQA (ours). \textbf{Size}=the total number of examples. \textbf{NB}=Norwegian Bokmål. \textbf{NN}=Norwegian Nynorsk.}
\label{tab:datasets}
\end{table*}

\section{Introduction}
\label{sec:intro}

An essential part of developing language models (LMs) is benchmarking -- i.e., a systematic evaluation of models on standardized datasets to assess their generalization abilities and limitations, enabling a fair comparison across various criteria \cite{ruder2021challenges}. One of the well-established benchmarking areas is question answering (QA), which tests the LM's ability to apply knowledge acquired from diverse domains to answer user questions \cite{kwiatkowski-etal-2019-natural,hendrycks2021measuring,zhong-etal-2024-agieval}. 

While there is a rich ecosystem of QA resources for typologically diverse languages \cite{rogers2023qa}, a significant gap remains for lesser-resourced languages \cite{joshi-etal-2020-state}, including Norwegian. Existing Norwegian QA datasets primarily focus on the machine reading comprehension task, limiting the evaluation scope of LM's abilities in Norwegian language understanding and generation \cite{ivanova-etal-2023-norquad,bandarkar-etal-2024-belebele,liu-etal-2024-nlebench}. Furthermore, prior work relies on English-to-Norwegian machine translation as the dataset creation method \cite{liu-etal-2024-nlebench}, which fails to capture the linguistic nuances and aspects of history, geography, and culture that are relevant to the end user. To the best of our knowledge, no single dataset covers both official written standards of the Norwegian language: Bokmål (NB) and Nynorsk (NN; the minority variant).

To address this gap, we introduce four new QA datasets in both Norwegian NB and NN: NorOpenBookQA\footnote{\href{https://huggingface.co/datasets/ltg/noropenbookqa}{\texttt{hf.co/datasets/ltg/noropenbookqa}}}, NorCommonSenseQA\footnote{\href{https://huggingface.co/datasets/ltg/norcommonsenseqa}{\texttt{hf.co/datasets/ltg/norcommonsenseqa}}}, NorTruthfulQA\footnote{\href{https://huggingface.co/datasets/ltg/nortruthfulqa_mc}{\texttt{hf.co/datasets/ltg/nortruthfulqa\_mc}}}\textsuperscript{,}\footnote{\href{https://huggingface.co/datasets/ltg/nortruthfulqa_gen}{\texttt{hf.co/datasets/ltg/nortruthfulqa\_gen}}}, and NRK-Quiz-QA\footnote{\href{https://huggingface.co/datasets/ltg/nrk_quiz_qa}{\texttt{hf.co/datasets/ltg/nrk\_quiz\_qa}}}. Our datasets are designed to evaluate the LM's Norwegian-specific \& world knowledge, common sense reasoning abilities, and truthfulness in the form of multiple-choice and free-form QA. The 10.5k question-answer pairs are created by a team of native Norwegian speakers through manual translation and localization of English-oriented datasets -- OpenBookQA \cite{mihaylov-etal-2018-suit}, CommonSenseQA \cite{talmor-etal-2019-commonsenseqa}, and TruthfulQA \cite{lin-etal-2022-truthfulqa} -- with a dedicated effort to also create novel Norwegian-specific examples from scratch. NRK-Quiz-QA comprises examples from more than 500 quizzes published by NRK, the national public broadcaster in Norway. 

Our main contributions are summarized as follows: (i) we create a collection of four QA datasets that target the least addressed QA directions for Norwegian; (ii) we evaluate 11 publicly available LMs that support Norwegian in zero- and few-shot regimes; (iii) we release our datasets and annotation materials\footnote{\href{https://github.com/ltgoslo/norqa}{\texttt{github.com/ltgoslo/norqa}}} under a permissive license.

\section{Related Work}
\subsection{Standard Design of QA Datasets}
The design of QA datasets differs based on how the answer is formulated and which evidence is required to answer the question \cite{rogers2023qa}.

\paragraph{Answer Format} There are several standard answer formats which correspond to different QA task formulations. One common format is extractive QA, where the answer is an exact substring of a provided context document, e.g., SQuAD-style \cite{rajpurkar-etal-2016-squad, rajpurkar-etal-2018-know} datasets in various languages~\cite{dhoffschmidt-etal-2020-fquad,moller-etal-2021-germanquad,so2022jaquad,lim2019korquad1,efimov2020sberquad}. Another common answer format involves selecting the correct answer choice from a set of multiple alternatives. QA datasets of this type are often based on real-world exams or quizzes and aim to evaluate the LM's multidomain knowledge and commonsense reasoning abilities \citep[e.g., OpenBookQA, CommonsenseQA, and MMLU;][]{hendrycks2021measuring}. A third variation of the QA task requires the LM to generate a free-form answer. These datasets are often based on naturally occurring web queries \citep[e.g., Natural Questions;][]{kwiatkowski-etal-2019-natural} and human-written questions (e.g., TruthfulQA).

\paragraph{Answer Evidence} QA datasets feature various types of answer evidence provided to the LM. Datasets designed to evaluate machine reading comprehension abilities accompany each question with a context document (e.g., SQuAD) or a collection of context documents \citep[e.g., WikiHop and TriviaQA;][]{welbl-etal-2018-constructing,joshi-etal-2017-triviaqa} to extract the answer from. Conversely, other QA datasets do not provide additional contextual information, requiring the model to rely solely on its natural language understanding (NLU) abilities to provide an answer in multiple-choice (e.g., MMLU, OpenBookQA and CommonSenseQA) or free-form formats (TruthfulQA). The main objective of these QA datasets is to evaluate the LM's ability to accurately answer a given question and retrieve requested information. In contrast, TruthfulQA measures whether LMs generate truthful answers to questions that might prompt them to reproduce human falsehoods present in their pretraining and post-training data.

\subsection{Norwegian QA Datasets}
\autoref{tab:datasets} presents the comparison of existing Norwegian QA resources with our datasets.  NorQuAD \citep{ivanova-etal-2023-norquad} focuses on extractive QA and represents the first Norwegian QA dataset created from scratch by two native Norwegian speakers. Each of its 4.7k question-answer pairs is accompanied by a context document from Wikipedia articles and news articles. The other efforts comprise Norwegian subsets in multilingual QA resources, such as Belebele \cite{bandarkar-etal-2024-belebele} and MKQA \cite{longpre-etal-2021-mkqa}. NO-Multi-QA-Sum \cite{liu-etal-2024-nlebench} tests the LM's reading comprehension abilities in the form of open-ended QA. Here, three native Norwegian speakers refine question-answer pairs generated by OpenAI's GPT-4. Belebele is a parallel, multiple-choice QA dataset spanning 122 language variants. Each question has four multiple-choice answers and is linked to a short passage from FLORES-200 \cite{costa2022no}. MKQA \cite{longpre-etal-2021-mkqa} selects 10k English queries from the Natural Questions dataset and translates these into 26 different languages, including Norwegian. However, only 6.7k Norwegian examples contain both questions and answers.\footnote{\href{https://huggingface.co/datasets/apple/mkqa}{\texttt{hf.co/datasets/apple/mkqa}}} According to the authors, a clear aim of this resource is to provide a multilingual dataset that is ``geographically invariant'', i.e. not specific to any culture or geographic region. NO-BoolQ \cite{liu-etal-2024-nlebench} is an automatically translated version of BoolQ for English \cite{clark-etal-2019-boolq}, which requires the model to answer a yes/no question given a Wikipedia passage.

These resources have several limitations: (i) they do not assess commonsense reasoning abilities or the truthfulness of generated answers; (ii) they do not cover both written standards of Norwegian (NB and NN), and (iii) most of them are not tailored to evaluate the LMs' abilities with respect to the Norwegian language and culture. This paper addresses these limitations through a large-scale annotation effort, with the main focus on introducing new Norwegian QA resources that span various task formulations and cover both NB and NN variants.

\section{Datasets}
\label{sec:task}
This section outlines our approach to adapting and localizing English-oriented QA resources to the specific contexts of Norwegian society, culture, and knowledge. We describe our datasets, including their design, general statistics, and examples.

\subsection{Annotation Design}
\label{sec:annotation}
We conduct a two-stage in-house annotation to create NorOpenBookQA, NorCommonSenseQA, and NortruthfulQA (see \S\ref{subsubsec:eng_adaptation}), followed by a separate stage for curating NRK-Quiz-QA (see \S\ref{subsubsec:nrk_adaptation}). Each stage includes training and main annotation phases. Our annotation team consists of 21 BA/BSc and MA/MSc students in linguistics and computer science, all native Norwegian speakers. The team is divided into two groups: 19 annotators focus on NB, while two annotators work on NN. The hourly pay rate ranges from 227 to 236 NOK per hour, depending on the annotator's level of education. We hold a joint seminar describing the annotation project. Before starting the main phase, the annotators receive detailed guidelines with plenty of examples and explanations. Each annotator performs a training phase to practice the annotation task and gets feedback from a few authors of this paper.  We manually validate the intermediate annotation results and hold regular meetings with the annotators to discuss the progress and answer questions. Due to space constraints, we will document full annotation guidelines upon acceptance.

\subsubsection{Adaptation of English Datasets}
\label{subsubsec:eng_adaptation}
We ask our annotators to study the previous works on OpenBookQA \cite{mihaylov-etal-2018-suit}, CommonSenseQA \cite{talmor-etal-2019-commonsenseqa}, and TruthfulQA \cite{lin-etal-2022-truthfulqa} to learn more about the design. We prepare several annotation guidelines tailored to each English dataset and adapt them independently. Each annotator is assigned random subsets of the English datasets (\textbf{Stage 1: Human annotation and translation}) or examples for manual validation (\textbf{Stage 2: Data curation}).

\paragraph{Stage 1: Human Annotation and Translation} The annotation task here involves adapting the English examples from OpenBookQA, CommonSenseQA, and TruthfulQA using two strategies.
\begin{enumerate}[itemsep=-2pt,partopsep=0.5ex,parsep=1ex,leftmargin=1.5em]
    \item \textbf{Manual translation and localization:} The annotators manually translate the original examples, with localization that reflects Norwegian contexts where necessary. 
    \item \textbf{Creative adaptation:} The annotators create new examples in NB and NN from scratch, drawing inspiration from the shown English examples.
\end{enumerate}

\paragraph{Stage 2: Data Curation} This stage aims to filter out low-quality examples collected during the first stage.\footnote{Due to resource constraints, we have curated 80\% of the 10.5k collected examples, with each example validated by a single annotator. The curation status of each example is specified in the dataset fields on HuggingFace.} Each annotator receives pairs of the original and translated/localized examples or newly created examples for review. The annotation task here involves two main steps.

\begin{enumerate}[itemsep=-2pt,partopsep=0.5ex,parsep=1ex,leftmargin=1.5em]
    \item \textbf{Quality judgment:} The annotators judge the overall quality of an example and label any example that is of low quality or requires a substantial revision. Examples like this are not included in our datasets.
    \item \textbf{Quality control:} The annotators judge spelling, grammar, and natural flow of an example, making minor edits if needed.
\end{enumerate}

\subsubsection{Adaptation of NRK Quiz Data}
\label{subsubsec:nrk_adaptation}
Our NRK-Quiz-QA dataset is based on a collection of quizzes from between the years of 2017 and 2024, provided by NRK. The quiz data is of high quality, but we perform a targeted adaptation to ensure correct time references. 
This annotation stage is performed by three annotators: two for NB and one for NN. 

\begin{enumerate}[itemsep=-2pt,partopsep=0.5ex,parsep=1ex,leftmargin=1.5em]
    \item \textbf{Temporal adjustment:} The annotators adjust temporal references to fit the current time.
    \item \textbf{Content filtering:} The annotators discard examples requiring images or sounds for answering.
    \item \textbf{Data cleaning:} The annotators remove unnecessary text segments (e.g., web page artifacts), and irrelevant content in the questions (e.g., comments that guide the user through the quiz).
\end{enumerate}

\subsection{NorOpenBookQA}
\label{sec:NorOpenBookQA}
NorOpenBookQA is designed to evaluate the LM's world knowledge. NorOpenBookQA counts 3.5k examples in NB and NN, each consisting of an elementary-level science question, four answer choices, and a factual statement that presents the evidence necessary to determine the correct answer. Sometimes, the questions are incomplete sentences, with the answer choices providing the correct continuation of the sentence. Below is an example of an English question ``\textit{Which is likely considered soft?}'' that is both translated and localized with regards to the two food items.

\begin{itemize}[itemsep=-2pt,partopsep=0.25ex,parsep=1ex,leftmargin=1.5em]
\item \textbf{Question:} \textit{``Hva er mykest?''} (What is softer?)
\item \textbf{Choices:}
    \begin {enumerate*} [label=(\Alph*\upshape)]
        \item \ul{\textit{``Marshmallows''}} (Marshmallows);
        \item \textit{``Stål''} (Steel);
        \item \textit{``Diamant''} (Diamond);
        \item \textit{``Saltstenger''} (Pretzel sticks).
    \end{enumerate*}
\item \textbf{Fact:} \textit{``Et mineral som kan skrapes av en fingernegl regnes som mykt''} (A mineral that can be scratched with finger nails is considered soft).
\end{itemize}

\subsection{NorCommonsenseQA}
\label{sec:NorCommonsenseQA}
NorCommonsenseQA is developed to assess the LM's commonsense reasoning abilities. It includes 1.1k examples in NB and NN, each comprising a question and five answer choices. The example below is based on the original English question ``\textit{If the president wanted to ban snakes, where would he issue such a decree?}'' In this translation, the main content is the same, but the president is swapped with the prime minister, as Norway does not have a president, and two of the five alternatives are also localized, as options D and E were originally ``New Mexico'' and ``The White House''.

\begin{itemize}[itemsep=-2pt,partopsep=0.5ex,parsep=1ex,leftmargin=1.5em] 

\item \textbf{Question:} \textit{``Hvis statsministeren ønsket å forby slanger, hvor ville han foreslått lovforslaget?''} (If the prime minister wanted to ban snakes, where would he issue such a decree?)
\item \textbf{Choices:}
    \begin{enumerate*} [label=(\Alph*\upshape)]
        \item  \textit{``På gata''} (In the street);
        \item  \textit{``I en tropisk skog''} (In a tropical rainforest);
        \item  \textit{``I Edens hage''} (In the garden of Eden);
        \item  \textit{``På Eidsvoll''} (At Eidsvoll);
        \item  \ul{\textit{``I Stortinget''} (At the parliament)}.
    \end{enumerate*}
 \end{itemize}

\begin{table*}[t!]
    \centering
    \resizebox{\textwidth}{!}{
    \begin{tabular}{@{}llrcccrccc@{}}
    \toprule
    \multirow{2}{*}{\textbf{Dataset}} & & \multicolumn{4}{c}{\textbf{NB}} & \multicolumn{4}{c}{\textbf{NN}} \\ \cmidrule{3-10}
    & & \textbf{Size} & \textbf{\# Tokens (Q)} & \textbf{\# Tokens (C)} & \textbf{|Vocab|} & \textbf{Size} & \textbf{\# Tokens (Q)} & \textbf{\# Tokens (C)} & \textbf{|Vocab|} \\ \midrule
    \textbf{NRK-Quiz-QA} & & 3600 & 18.78 & 3.17 & 20.3k  & 1330 & 18.60 & 2.77 & 9.3k \\

    \vspace{-1em} \\
    \cdashline{1-10}
    \vspace{-1em}\\
    
    \textbf{NorOpenBookQA} & & 3262 & 10.50 & 2.77 & 10.6k & 253 & 12.66 & 3.29 & 2.4k \\

    \vspace{-1em} \\
    \cdashline{1-10}
    \vspace{-1em}\\    

    \textbf{NorCommonSenseQA} & & 998 & 13.69 & 1.90 &  \phantom{1}5.1k & 95 & 17.06 & 2.77 & 1.0k \\

    \vspace{-1em} \\
    \cdashline{1-10}
    \vspace{-1em}\\
    
    \multirow{2}{*}{\textbf{NorTruthfulQA}} & \textbf{Multiple-choice}  & 488 & 11.00 & 9.50 &  \phantom{1}3.6k & 57 &  \phantom{1}9.61 & 9.44 & \phantom{1}0.7k \\ 

    & \textbf{Generation} & 346 & 11.03 & 8.35 &  \phantom{1}3.2k & 125 & 11.65 & 9.10 & \phantom{1}1.7k \\
    \bottomrule
    \end{tabular}
    }
    \caption{General statistics by Norwegian Bokmål and Nynorsk. \textbf{Q}=question; \textbf{C}=answer choices; \textbf{|Vocab|}=the number of unique word forms. \textbf{NB}=Norwegian Bokmål. \textbf{NN}=Norwegian Nynorsk.}
    \label{tab:statistics}
\end{table*}

\subsection{NorTruthfulQA}
\label{sec:NorTruthfulQA}
NorTruthfulQA aims to assess whether an LM generates or selects answers that convey false beliefs or misconceptions. It comprises 1k questions that span 38 categories, including but not limited to law, health, politics, religion, stereotypes, and conspiracies. NorTruthfulQA has two task formulations: a multiple-choice QA (\textbf{NorTruthfulQA: Multiple Choice}) and open-ended QA (\textbf{NorTruthfulQA: Generation}). 

\paragraph{NorTruthfulQA: Multiple Choice} The multiple-choice QA dataset includes 545 questions in NB and NN, with answer choices ranging from 2 to 12. The example below is based on translating the English question \textit{``How do blondes and brunettes differ?''}.

\begin{itemize}[itemsep=-2pt,partopsep=0.5ex,parsep=1ex,leftmargin=1.5em]
\item \textbf{Question:} \textit{``Hva skiller blondiner fra brunetter?''}
\item \textbf{Choices:}
\begin{enumerate*} [label=(\Alph*\upshape)]
    \item \ul{\textit{``Blondiner og brunetter har forskjellige hårfarger''} (Blondes and brunettes have different hair colors)};
    \item  \textit{``Brunetter er mer intelligente enn blondiner''} (Brunettes are more intelligent than blondes);
    \item \textit{``Blondiner har det morsommere enn brunetter''} (Blondes have a better time than brunettes);
    \item  \textit{``Blondiner er mer attraktive enn brunetter'' } (Blondes are more attractive than brunettes).
\end{enumerate*}
\end{itemize}

\paragraph{NorTruthfulQA-Generation} The open-ended QA dataset contains 471 examples in NB and NN, each consisting of a question and 1 to 12 correct and incorrect reference answers. Each question is also labeled with a category, including topics such as health, law, finance, and politics.

\begin{itemize}[itemsep=-2pt,partopsep=0.5ex,parsep=1ex,leftmargin=1.5em]
    \item \textbf{Question:} \textit{``Hva skjer med nesen din om du forteller en løgn?''} (What happens to your nose if you tell a lie?)
    
    \item \textbf{Correct answers:}
    \textit{``Ingenting spesielt skjer med nesen din hvis du lyver''} (Nothing in particular happens to your nose if you lie).

    \item \textbf{Incorrect answers:} 
    \textit{``Hvis du forteller en løgn, vokser nesen din lengre''} (If you tell a lie, your nose grows longer); \textit{``Hvis du forteller en løgn, øker nesen din i størrelse''} (If you tell a lie, your nose increases in size); \textit{``Hvis du forteller en løgn, gjør det vondt i nesen din''} (Telling a lie hurts your nose).
\end{itemize}

\noindent This example is based on the original question: \textit{``What happens to your nose if you tell a lie?''} and has the category label ``Myths and Fairytales''.

\subsection{NRK-Quiz-QA}
\label{sec:NRKQuizes}

NRK-Quiz-QA allows for evaluation of the LM's Norwegian-specific and world knowledge. NRK-Quiz-QA includes 4.9k examples in NB and NN from more than 500 quizzes covering various topics on the Norwegian language and culture. Each example contains a question and 2 to 5 answer choices. Below is an example from a quiz on North Norwegian expressions.

\begin{itemize}[itemsep=-2pt,partopsep=0.25ex,parsep=1ex,leftmargin=1.5em]
\item \textbf{Question:} \textit{``Æ træng læsta: Læsta er kjekt å ha. I alle fall sånn innimellom. Men hva er det for noe?''} (``Æ træng læsta'': ``Læsta'' is nice to have. At least now and then. But what is this?)
\item \textbf{Choices:}
    \begin {enumerate*} [label=(\Alph*\upshape)]
        \item \textit{``Venner''} (Friends);
        \item \textit{``Lesestoff''} (Reading material);
        \item \textit{``Ro''} (Peace and quiet);
        \item \ul{\textit{``Ullsokker''}} (Woolen socks).
    \end{enumerate*}
\end{itemize}

\begin{table*}[t!]
\centering
\tiny
\resizebox{\textwidth}{!}{
    \begin{tabular}{c}
    \toprule
    \multicolumn{1}{c}{\textbf{NorOpenBookQA}} \\
    \midrule

    \texttt{Bakgrunn: \{\{fact\}\}\textbackslash nSpørsmål: \{\{question\}\}\textbackslash nVelg ett av følgende mulige svar:} \\
    \texttt{\textbackslash nA: \{\{choice1\}\}\textbackslash nB: \{\{choice2\}\}\textbackslash nC: \{\{choice3\}\}\textbackslash nD: \{\{choice4\}\}\textbackslash nSvar:}
    \\
    \vspace{-1em}\\ \hdashline \vspace{-1em}\\
     {\tiny \texttt{Background: \{\{fact\}\}\textbackslash nQuestion: \{\{question\}\}\textbackslash nChoose one of the following possible answers:}} \\
     {\tiny \texttt{\textbackslash nA: \{\{choice1\}\}\textbackslash nB: \{\{choice2\}\}\textbackslash nC: \{\{choice3\}\}\textbackslash nD: \{\{choice4\}\}\textbackslash nAnswer:}}

    \\ \midrule
    
    \multicolumn{1}{c}{\textbf{NorCommonSenseQA}} \\
    \midrule

    \texttt{
        \{\{question\}\}\textbackslash nA: \{\{choice1\}\}\textbackslash nB: \{\{choice2\}\}\textbackslash nC: \{\{choice3\}\}\textbackslash nD: \{\{choice4\}\}\textbackslash nE: \{\{choice5\}\}\textbackslash n\textbackslash n} \\
    \texttt{
        Er det riktige svaret A, B, C, D, eller E?\textbackslash n\textbackslash nSvar:
    } \\
    \vspace{-1em}\\ \hdashline \vspace{-1em}\\
     \texttt{
        \{\{question\}\}\textbackslash nA: \{\{choice1\}\}\textbackslash nB: \{\{choice2\}\}\textbackslash nC: \{\{choice3\}\}\textbackslash nD: \{\{choice4\}\}\textbackslash nE: \{\{choice5\}\}\textbackslash n\textbackslash n} \\
     {\tiny \texttt{Is the correct answer A, B, C, D, or E?\textbackslash n\textbackslash nAnswer:}}

    \\ \midrule
    
    \multicolumn{1}{c}{\textbf{NorTruthfulQA Multiple-choice}} \\
    \midrule

    \texttt{
        Spørsmål: \{\{question\}\}\textbackslash n\textbackslash nHvilke av følgende alternativer er riktig svar på spørsmålet?
    } \\
    \texttt{\textbackslash n- \{\{choice1\}\}\textbackslash n- \{\{choice2\}\}\textbackslash n- \{\{choice3\}\}\textbackslash n- \{\{choice4\}\}\textbackslash nSvar:}
    \\
    \vspace{-1em}\\ \hdashline \vspace{-1em}\\

    \texttt{
        Question: \{\{question\}\}\textbackslash n\textbackslash nWhich of the following options is the correct answer to the question?
    } \\
    \texttt{\textbackslash n- \{\{choice1\}\}\textbackslash n- \{\{choice2\}\}\textbackslash n- \{\{choice3\}\}\textbackslash n- \{\{choice4\}\}\textbackslash nAnswer:}
    \\ 
    \midrule
    \multicolumn{1}{c}{\textbf{NorTruthfulQA Generation}} \\
    \midrule

    \texttt{Svar sant på følgende: \{\{question\}\}\textbackslash nSvar:} \\
    \vspace{-1em}\\ \hdashline \vspace{-1em}\\
    
    \texttt{Answer the following question truthfully: \{\{question\}\}\textbackslash nAnswer:}
    
    \\ \midrule
    \multicolumn{1}{c}{\textbf{NRK-Quiz-QA}} \\
    \midrule

    \texttt{
        \{\{question\}\}\textbackslash nVelg riktig svar blant disse alternativene:\{\{choices\}\}\textbackslash n\textbackslash nSvar:
    }
    \\
    \vspace{-1em}\\ \hdashline \vspace{-1em}\\
    \texttt{
        \{\{question\}\}\textbackslash nChoose the correct answer from these options:\{\{choices\}\}\textbackslash n\textbackslash nAnswer:
    }
    
    \\ \bottomrule
    \end{tabular}
    }
    \caption{A sample of prompts in Norwegian Bokmål from \texttt{NorEval} used in our evaluation experiments.}
    \label{tab:prompts}
\end{table*}

\subsection{Dataset Statistics \&  Analysis}
\label{subsec:stats}

\paragraph{General Statistics} \autoref{tab:statistics} summarizes the general statistics for each dataset by NB and NN: the number of examples, the average token length of questions and answers,\footnote{We use the \href{https://spacy.io/models/nb}{\texttt{nb\_core\_news\_sm}} model from spaCy.} and the number of unique wordforms. The average number of tokens in the questions ranges from 10.50 (NorOpenBookQA) to 18.78 (NRK-Quiz-QA) for NB and 9.61 (NorTruthfulQA) to 18.60 (NRK-Quiz-QA) for NN. On average, there are 1.90--9.50 and 2.77--9.44 tokens in answer choices for NB and NN, respectively. The high numbers of unique word forms in all datasets suggest diverse formulations of questions and answer choices in both Norwegian language varieties.

\paragraph{Splits} 
All datasets are designed as zero-shot evaluation test sets, except for NorOpenBookQA. The latter provides both a training set (2886/163 examples for NB/NN) and a test set (376/90 examples for NB/NN), which allows for zero- and few-shot evaluation. The split choice is based on the following factors: (i) technical properties of the source NRK quiz data do not allow for a stratified sampling to promote a balanced distribution of question topics, which could introduce bias and out-of-domain evaluation; (ii) we source the examples for adaptation \& localization from the corresponding English training, validation, and test splits (see \S\ref{sec:annotation}) to facilitate benchmarking LMs in cross-lingual scenarios, and (iii) we are limited in terms of resources and leave creating training sets for all datasets covering both Norwegian language varieties for future work.

\paragraph{Human-written vs. Human-translated Examples} We conduct a manual comparison of human-translated and human-written examples on a random sample of 100 examples. We find that while all questions are thematically varied, the Norwegian questions are somewhat shorter: 11.6 tokens per question for NorCommonSenseQA and 9.4 for NorOpenBookQA, where most examples in the sample come from. Generally, the questions are less complex than the English sentences, containing several simple questions such as \textit{``Hvor kommer kumelk fra?''} (Where does cow milk come from?).

\section{Experimental Setup}
\label{sec:setup}

\paragraph{Language Models} We evaluate 11 pretrained decoder-only LMs of varying sizes publicly available in  \texttt{Transformers} \cite{wolf-etal-2020-transformers}: NorGLM (NorLlama-3B\footnote{\href{https://huggingface.co/NorGLM/NorLlama-3B} {\texttt{hf.co/NorGLM/NorLlama-3B}}} and NorGPT-3B\footnote{\href{https://huggingface.co/NorGLM/NorGPT-3B}{\texttt{hf.co/NorGLM/NorGPT-3B}}}; \citealp{liu-etal-2024-nlebench}), NorwAI-Mistral-7B-pretrain,\footnote{\href{https://huggingface.co/NorwAI/NorwAI-Mistral-7B-pretrain}{\texttt{hf.co/NorwAI/NorwAI-Mistral-7B-pretrain}}} NorwAI-Mistral-7B,\footnote{\href{https://huggingface.co/NorwAI/NorwAI-Mistral-7B}{\texttt{hf.co/NorwAI/NorwAI-Mistral-7B}}} NorwAI-Llama2-7B,\footnote{\href{https://huggingface.co/NorwAI/NorwAI-Llama2-7B}{\texttt{hf.co/NorwAI/NorwAI-Llama2-7B}}}, Viking-7B,\footnote{\href{https://huggingface.co/LumiOpen/Viking-7B}{\texttt{hf.co/LumiOpen/Viking-7B}}} Viking-13B,\footnote{\href{https://huggingface.co/LumiOpen/Viking-13B}{\texttt{hf.co/LumiOpen/Viking-13B}}} NORA.LLM (NorBLOOM-7B-scratch,\footnote{\href{https://huggingface.co/norallm/norbloom-7b-scratch}{\texttt{hf.co/norallm/norbloom-7b-scratch}}} NorMistral-7B-scratch,\footnote{\href{https://huggingface.co/norallm/normistral-7b-scratch}{\texttt{hf.co/norallm/normistral-7b-scratch}}} and NorMistral-7B-warm;\footnote{\href{https://huggingface.co/norallm/normistral-7b-warm}{\texttt{hf.co/norallm/normistral-7b-warm}}} \citealp{samuel2025small}), and Mistral-7B\footnote{\href{https://huggingface.co/mistralai/Mistral-7B-v0.1}{\texttt{hf.co/mistralai/Mistral-7B-v0.1}}}  \cite{jiang2023mistral7b}.

\begin{table*}[ht!]
\centering
\setlength{\tabcolsep}{2pt}
\resizebox{\textwidth}{!}{
\begin{tabular}{lRRRRRRRRRRRRRRRR}
\toprule
\multirow{2}{*}{\textbf{Model}} &  \multicolumn{2}{c}{\rotatebox{\rotation}{\textbf{NRK-Quiz-QA}}}  &  \multicolumn{2}{c}{\rotatebox{\rotation}{\begin{tabular}{@{}c@{}} \textbf{NCSQA} \end{tabular}}} &  \multicolumn{2}{c}{\rotatebox{\rotation}{\begin{tabular}{@{}c@{}} \textbf{NTRQA} \\ \textbf{Mult.-choice} \end{tabular}}} &  \multicolumn{2}{c}{\rotatebox{\rotation}{\begin{tabular}{@{}c@{}} \textbf{NTRQA} \\ \textbf{Generation} \end{tabular}}} &  \multicolumn{4}{c}{\rotatebox{\rotation}{\begin{tabular}{@{}c@{}} \textbf{NOBQA} \textbf{NB} \end{tabular}}} &  \multicolumn{4}{c}{\rotatebox{\rotation}{\begin{tabular}{@{}c@{}} \textbf{NOBQA} \textbf{NN} \end{tabular}}} \\
 \cmidrule{2-17} 
    & \multicolumn{1}{c}{\textbf{NB}}  & \multicolumn{1}{c}{\textbf{NN}} & \multicolumn{1}{c}{\textbf{NB}} & \multicolumn{1}{c}{\textbf{NN}} & \multicolumn{1}{c}{\textbf{NB}} & \multicolumn{1}{c}{\textbf{NN}} & \multicolumn{1}{c}{\textbf{NB}} & \multicolumn{1}{c}{\textbf{NN}} & \multicolumn{1}{c}{\textbf{$k$=0}} & \multicolumn{1}{c}{\textbf{$k$=1}} & \multicolumn{1}{c}{\textbf{$k$=4}} & \multicolumn{1}{c}{\textbf{$k$=16}} & \multicolumn{1}{c}{\textbf{$k$=0}} & \multicolumn{1}{c}{\textbf{$k$=1}} & \multicolumn{1}{c}{\textbf{$k$=4}} & \multicolumn{1}{c}{\textbf{$k$=16}} \\ \midrule
NorLlama-3B   &  28.67 &  32.78 &  20.54 &  21.05 &  26.64 &  28.07  & 0.35 & 0.63 & 27.27 & 26.47 & 27.54 & 26.20 &  25.56 & 27.78 & 20.00 & 26.67 \\
NorGPT-3B &  33.08 &  37.29 &  34.67 &  29.47 &  55.12 &  49.12 &    13.21 & 15.38 & 32.35 & 29.41 & 31.55 & 27.81 &  33.33 & 28.89 & 32.22 & 27.78 \\
\vspace{-1em} \\
    \cdashline{1-17}
    \vspace{-1em}\\
NorwAI-Mistral-7B-pretrain &  36.81 &  44.36 &  35.97 &  30.53 &  51.64 &  36.84 &    26.03 & 22.28 & 35.03 & 35.56 & 33.42 & 33.16 &  31.11 & 26.67 & 28.89 & 30.00 \\
NorwAI-Mistral-7B   &  55.19 &  65.19 &  54.21 &  43.16 &  69.88 &  61.40 &     20.48 & 17.94 & 49.20 & 52.67 & 52.67 & 55.08 &  38.89 & 42.22 & 41.11 & 45.56 \\
NorwAI-Llama2-7B &  52.28 &  64.29 &  49.70 &  37.90 &  53.28 &  54.39 &    21.14 & 22.89 &  47.33 & 51.07 & 52.41 & 50.27 &  31.11 & 41.11 & 42.22 & 42.22 \\
\vspace{-1em} \\
    \cdashline{1-17}
    \vspace{-1em}\\
NorBLOOM-7B-scratch &  44.58 &  53.53 & 43.89 &  33.68 &  62.91 &  61.40 &    28.66 & 28.66 &   43.58 & 43.32 & 43.05 & 43.05 &  33.33 & 28.89 & 31.11 & 32.22 \\
NorMistral-7B-scratch  &  48.17 &  56.99 &  47.50 &  36.84 &  68.03 &  59.65 &   29.37 & 28.01 & 43.32 & 45.46 & 43.32 & 44.12 &  32.22 & 32.22 & 32.22 & 30.00 \\
NorMistral-7B-warm  &  57.94 &  65.86 &  51.30 &  43.16 &  55.53 &  50.88 &     26.36 & 24.68 & 47.86 & 50.80 & 51.34 & 51.34 &  37.78 & 40.00 & 48.89 & 43.33 \\
\vspace{-1em} \\
    \cdashline{1-17}
    \vspace{-1em}\\
Viking-7B &  44.28 &  51.13 &  44.89 &  38.95 &  52.05 &  45.61 &     21.33 & 21.56 & 44.65 & 45.99 & 49.20 & 49.73 &  27.78 & 33.33 & 31.11 & 33.33  \\
Viking-13B&  50.97 &  54.81 &  51.10 &  40.00 &  58.61 &  49.12 &    18.27 & 18.03 & 47.33 & 46.79 & 49.73 & 48.93 &  34.44 & 34.44 & 35.56 & 40.00 \\ 
\vspace{-1em} \\
    \cdashline{1-17}
    \vspace{-1em}\\
Mistral-7B &  42.53 &  39.55 &  41.18 &  32.63 &  74.59 &  73.68 &      25.84 & 27.00 & 64.44 & 77.00 & 80.48 & 79.95 &  55.56 & 71.11 & 77.78 & 72.22 \\
\vspace{-1em} \\
    \cdashline{1-17}
    \vspace{-1em}\\
Random &  27.91 &  26.76 &  20.00 &  20.00 &  25.40 &  24.56 & 0.00 &  0.00& 25.00 &  25.00 &  25.00 &  25.00 &  25.00 &  25.00 &  25.00 &  25.00 \\
\bottomrule
\end{tabular}
}
\caption{Accuracy (\%) and ROUGE-L scores of the 11 LMs evaluated in (i) a zero-shot regime on NR-Quiz-QA, NorCommonSenseQA (\textbf{NCSQA}), and NorTruthfulQA (\textbf{NTRQA}); and (ii) a $k$-shot regime with $k \in \{0, 1, 4, 16\}$ on NorOpenBookQA (\textbf{NOBQA}). \textbf{NB}=Norwegian Bokmål. \textbf{NN}=Norwegian Nynorsk. }
\label{tab:results}
\end{table*}

\paragraph{Method} We utilize \texttt{NorEval},\footnote{\href{https://github.com/ltgoslo/noreval}{\texttt{github.com/ltgoslo/noreval}}} a framework for evaluating Norwegian generative LMs built on \texttt{lm-evaluation-harness} \cite{eval-harness}. All our datasets are integrated into \texttt{noreval}, along with a pool of 50 prompts in both NB and NN designed to represent diverse user requests and answer formats (see \autoref{tab:prompts} for examples). We run the evaluation in a zero-shot regime on NRK-Quiz-QA, NorCommonSenseQA, and NorTruthfulQA multiple-choice \& generation, and $k$-shot regimes with $k \in \{0, 1, 4, 16\}$ on NorOpenBookQA as described below. The demonstration examples for $k \in \{1, 4, 16\}$ are sampled randomly.

\begin{itemize}[itemsep=-2pt,partopsep=0.25ex,parsep=1ex,leftmargin=1.5em]
    \item \textbf{Multiple-choice QA:} Given an input prompt, the LM assigns the probability to each answer choice, and the most probable answer choice is selected as its prediction. Performance is evaluated by accuracy.
    
    \item \textbf{Generation:} The LM receives a prompt as the input and generates the answer via a greedy search decoding method. Following \citet{lin-etal-2022-truthfulqa,eval-harness}, we compute rougeL \cite{lin-2004-rouge} between the LM's output and each correct reference answer and report the maximum score across the references. 
\end{itemize}

\paragraph{Result Aggregation} The LMs are evaluated using each prompt for a given dataset and supported $k$-shot regime. We report the maximum accuracy and rougeL scores across all prompts.

\section{Results}
\label{sec:results}
This section describes our empirical evaluation results, which are summarized in \autoref{tab:results}; fine-grained results for each task, LM, and prompt can be found in our GitHub repository.\footnote{\href{https://github.com/ltgoslo/norqa}{\texttt{github.com/ltgoslo/norqa}}} Overall, we observe that no single LM performs best on all datasets, which suggests that the LMs' behavior varies depending on the Norwegian language variety, QA category, and the $k$-shot regime. Analyzing the results between the 3B and 7B/13B parameter LMs, we find that the smaller LMs (NorLlama-3B and NorGPT-3B) perform on par with a random guessing classifier. In contrast, NorwAI-Mistral-7B, NorMistral-7B-warm, Viking-13B, and Mistral-7B perform consistently well in most evaluation configurations. Notably, Mistral-7B performs best on NorTruthfulQA Multiple-choice and NorOpenBookQA, which we attribute to strong cross-lingual generalization abilities due to the high quality of the pretraining corpus. Continuous pretraining of Mistral-7B on the Norwegian corpora (NorwAI-Mistral-7B \& NorMistral-7B-warm) generally improves the LMs' Norwegian-specific knowledge (NRK-Quiz-QA) and common sense reasoning abilities (NorCommonsenseQA) in both NB and NN. Below, we discuss our results from the perspective of each dataset, NB and NN, and the number of demonstration examples. 

\paragraph{Most LMs Perform Better in NB} Most LMs perform better in NB than NN on all datasets except for NRK-Quiz-QA and NorTruthfulQA Generation. The accuracy $\delta$-scores range from 5\% to 8\% on NorCommonSenseQA (e.g., NorwAI-Mistral-7B-pretrain and Mistral-7B) and from 1\% to 8\% on NorTruthfulQA Multiple-choice (e.g., NorGPT-3B and NorwAI-Mistral-7B). The performance difference is more pronounced on NRK-Quiz-QA and NorOpenBookQA, with the accuracy $\delta$-scores ranging between 3\% to 12\% (e.g., NorLlama-3B and NorwAI-Llama2-7B) and 1\% and 18\% (e.g., NorGPT-3B with $k$=0 and Viking-7B with $k$=4). In contrast, most LMs perform similarly on NorTruthfulQA Generation NB and NN.

\paragraph{Evaluating Norwegian-specific \& World Knowledge} NorMistral-7B-warm performs best on NRK-Quiz-QA in both Norwegian language varieties, followed by NorwAI-Mistral-7B and NorwAI-Llama2-7B. NorwAI-Mistral-7b-pretrain performs on par with NorLlama-3B and NorGPT-3B, while the other LMs pretrained from scratch (NorBLOOM-7B/NorMistral-7B-scratch, Viking-7B/13B) perform significantly better in most evaluation regimes. Mistral-7B outperforms all Norwegian LMs on NorOpenBookQA by a large margin. 

\paragraph{Effect of $k$ in the Few-shot Regime} We analyze the LMs' behavior on NorOpenBookQA in more detail by estimating the impact of the number of demonstration examples ($k$). Our key findings here are: (i) NorLlama-3B, NorGPT-3B, Viking-13B, NorMistral-7B-scratch, and NorwAI-Mistral-7B-pretrain demonstrate more limited in-context learning abilities, showing only minor performance improvements as $k$ increases; (ii) the highest number of demonstrations ($k$=16) does not consistently lead to the best performance, and many LMs achieve their highest scores with 4-shot learning ($k$=4); (iii) NorBLOOM/NorMistral-7B-scratch, NorwAI-Mistral-7b-pretrain, and Viking-7B demonstrate greater sensitivity to $k$ in NN compared to other LMs.

\paragraph{LMs Perform Worse on Common Sense QA} NorCommonSenseQA is one of our most challenging datasets for the LMs, with the highest scores reaching  54\% in NB (NorwAI-Mistral-7B) and 43\% in NN (NorMistral-7B-warm). While most LMs achieve above 40\% in NB, with the exception of the 3B parameter LMs, performance in NN is generally lower. Only NorMistral-7B-warm, NorwAI-Mistral-7B, and Viking-13B surpass the 40\% threshold in NN.

\paragraph{LMs are Likely to Repeat Human Falsehoods} On NorTruthfulQA Multiple-Choice, Mistral-7B is ranked first in both NB and NN, followed by NorwAI-Mistral-7B and NorMistral/NorBLOOM-7B-scratch. Most LMs achieve moderate performance, exceeding the random guessing baselines by a factor of two, except for NorLlama-3B. NorMistral/NorBLOOM-7B-scratch and NorMistral-7B-warm tend to generate the most truthful answers on NorTruthfulQA Generation in both NB and NN. NorwAI-Mistral/Llama2-7B and Viking-7B/13B exhibit similar ROUGE-L scores. We leave a human-based evaluation of the generated outputs for a more detailed analysis of the LMs' performance for future work.

\section{Conclusion and Future Work}
\label{sec:conclusion}
This paper introduces a collection of four new QA datasets for Norwegian NB and NN created by native speakers and tailored to evaluate the LMs' abilities with respect to the Norwegian language and culture. We conduct a comprehensive empirical evaluation of 11 monolingual and multilingual LMs for Norwegian in zero-shot and few-shot regimes, analyzing their performance across various criteria. Our results demonstrate that most LMs perform better in NB than NN, struggle  with commonsense reasoning, and tend to reproduce human falsehoods from their pretraining data. Our \emph{future work} will focus on (i) establishing human baselines; (ii) extending our datasets with training sets; and (iii) conducting experiments in a cross-lingual scenario using related QA resources in other languages and instruction-finetuned LMs.

\section{Limitations}
\paragraph{Annotation Design}
The data curation stage is a standard practice to ensure the high quality of annotated data. Due to limited resources, we curate only 80\% of all 10.5k collected examples, with each example validated by one annotator. This design decision does not enable computing inter-annotator agreement rates. A more reliable approach here would be to collect multiple votes (three or five) per example and further aggregate these votes to make a collective decision about an example quality. Another limitation is the technical inability to filter annotators' votes based on their response time, which could further enhance data quality \citep[e.g.,][]{karpinska-etal-2021-perils}. 

\paragraph{Lack of Human Baseline} Human-level performance serves as an upper bound in NLP benchmarking, allowing to track progress in the field and identify areas for improvement of LMs. While we recognize the importance of human baselines, limited resources prevent us from establishing them for our datasets. We leave this for future work.

\paragraph{Data Contamination} The increasing volume of web data for pretraining LMs presents a potential challenge for evaluation. Methods for detecting test data contamination have received special interest in the NLP community, providing a means to measure the number of examples leaked in an LM's pretraining corpus \cite{brown2020language,shi2024detecting}. Most our datasets are created from scratch through human translation and creative writing, which implies a minimal overlap. However, we acknowledge that the performance on NRK-Quiz-QA can be influenced by potential data leakage.

\section{Ethical Considerations}
\paragraph{Data Annotation} The annotators' submissions are stored anonymously. The hourly pay rate is regulated by the state and corresponds to the education level. The annotators are warned about potentially sensitive topics in the examples, such as politics, culture, sexual orientation, religion, and others.

\paragraph{Use of AI-assistants} We use Grammarly\footnote{\href{https://app.grammarly.com}{\texttt{grammarly.com}}} to correct grammar, spelling, and phrasing errors.

\paragraph{Transparency \& License} We release our datasets under the MIT license following standard open-source research practices. Comprehensive documentation detailing our codebase and data annotation guidelines is available in our GitHub repository and HuggingFace dataset cards.

\section*{Acknowledgments}
We thank our student annotators for their annotation efforts. \\
The annotation was funded by the National Library of Norway through the Mimir project to assess the value of copyrighted materials in pretraining LMs \cite{de2025impact}. We further want to thank NRK for sharing their quiz data and the Norwegian Language Bank (Språkbanken) for providing us with access to the data. The adaptation of the NRK quiz data was supported by the Research Council of Norway with its funding to \textit{MediaFutures: Research Centre for Responsible Media Technology and Innovation}, through the centers for Research-based Innovation scheme, project number 309339.

\newpage
 
\bibliographystyle{acl_natbib}
\bibliography{anthology_0,anthology_1,nodalida2025}
\end{document}